\RequirePackage{fix-cm}
\documentclass[smallextended]{svjour3}       % onecolumn (second format)
\smartqed  % flush right qed marks, e.g. at end of proof
\usepackage{graphicx}
\usepackage{booktabs}
\usepackage{mathtools}
\usepackage{amssymb}
\usepackage{graphicx}
\usepackage{float}
\usepackage[ruled,vlined]{algorithm2e}
\usepackage[authoryear,longnamesfirst]{natbib}
\usepackage{xcolor}

\begin{document}

\title{Bayesian Optimization and Deep Learning for steering wheel angle prediction
}


\author{Alessandro Riboni \and
        Nicol\`o Ghioldi \and
        Antonio Candelieri \and
        Matteo Borrotti$^*$
}

%\authorrunning{A. Riboni et~al.} % if too long for running head

\institute{A. Riboni \at
              Department of Informatics, Systems and Communication, University of Milano-Bicocca, Italy \\
              \email{a.riboni2@campus.unimib.it}           %  \\
           \and
           N. Ghioldi \at
              Department of Economics, Management and Statistics, University of Milano-Bicocca, Italy \\
              \email{a.riboni2@campus.unimib.it}           %  \\
              \and
            A. Candelieri \at
              Department of Economics, Management and Statistics, University of Milano-Bicocca, Italy \\
              \email{antionio.candelieri@unimib.it}           %  \\
              \and
        M. Borrotti ($^*$Corresponding author) \at
              Department of Economics, Management and Statistics, University of Milano-Bicocca, Italy \\
              \email{matteo.borrotti@unimib.it}           %  \\
}

\date{Received: date / Accepted: date}

\maketitle

\begin{abstract}
Automated driving systems (ADS) have undergone a significant improvement in the last years. ADS and more precisely self-driving cars technologies will change the way we perceive and know the word of transportation systems in terms of user experience, mode choices and business models. The emerging field of Deep Learning (DL) has been successfully applied for the development of innovative ADS solutions. However, the attempt to single out the best deep neural network architecture and tuning its hyperparameters are all expensive processes, both in terms of time and computational resources. In this work, \textit{Bayesian Optimization} (BO) is used to optimize the hyperparameters of a  \textit{Spatiotemporal-Long Short Term Memory} (ST-LSTM) network with the aim to obtain an accurate model for the prediction of the steering angle in a ADS. BO was able to identify, within a limited number of trials, a model -- namely BO\_ST-LSTM -- which resulted, on a public dataset, the most accurate when compared to classical end-to-end driving models.
%\subclass{MSC code1 \and MSC code2 \and more}
\end{abstract}

\section{Introduction}

Over the last decade, significant progress has been made in automated driving systems (ADS). Given the current momentum and progress, ADS can be expected to continue to advance as variety of ADS products are going to become commercially available in the space of a decade \citep{Chan2017}. It is envisioned that automated driving technology will lead to a paradigm shift in transportation systems in terms of user experience, mode choices and business models. Nowadays, a greater number of industrialists are increasing their investments in self-driving cars technologies and, more generally, in the automotive sector. ADS research and an increasing number of industrial implementations have been catalyzed by the accumulated knowledge in vehicle dynamics in the wake of breakthroughs in computer vision caused by the advent of deep learning \citep{Krizhevsky2012, Bojarski2017, Kocic2019, Li2021} and the availability of new sensor modalities such as lidar \citep{Schwarz2010}. 

Deep Learning (DL) has been widely used for the implementation of ADSs. Starting from the work of \citet{Krizhevsky2012}, different DL approaches have been proposed. \citet{Bojarski2017} proposed a deep neural network (\textit{PilotNet}) for predicting the steering angles thanks to a set of images captured from the car. Based on \citet{Krizhevsky2012}, \citet{Kocic2019} proposed the so called \textit{J-net}, that is a DL model suitable for end-to-end systems. All of the above mentioned methods follow direct supervised training strategies. A review of these methods for ADS can be found in \citet{Yurtsever2019}. In this context, a ground truth is required for training, which normally consists in the ego-action sequence of an expert human driver, while the network learns to imitate the driver. More precisely, the training is performed on a set of camera images taken from the front car (input), which is a raw signal (\textit{i.e.} pixels) and, for example, the steering angles (output) used to control the car. Previously mentioned methods are then trained using road images paired with the steering angles generated by a human tasked with driving a data-collection car. The prediction tasks are solved using Convolutional Neural Networks (CNNs) \citep{Goodfellow2017}. Generally, CNNs use layers that convolve the inputs with filters and compress them. As a result, CNNs can find features and transform them into simpler representations. This ability has made them useful in many different applications (\textit{e.g.} arts and image classification, identification of movements) \citep{Balderas2019}. However, CNNs predict one frame at a time and generate future images recursively, which are prone to focus on spatial appearances and relatively weak in capturing long-term motions \citep{Wang2017}. In order to capture a temporal relation, a possible solution is the use of Long Short Term Memory (LSTM) \citep{Goodfellow2017} networks. The core idea behind the LSTM architecture is a memory cell which can maintain its state over time as non-linear gating units which regulate the information flow into and out of the cell \citep{Greff2017}. LSTM has been successfully used in different fields, ranging from weather prediction \citep{Farah2021} to brain tumor detection \citep{Amin2020}.

In order to overcome the limits of CNNs and exploit the advantages of LSTMs, \citet{Shi2015} proposed the convolutional LSTM (ConvLSTM) network, which is able to model the spatiotemporal structures simultaneously by explicitly encoding the spatial information into tensors, thus overcoming the limitation of vector-variate representations in standard LSTM where the spatial information is lost. Similar approaches have been also used in the context of ADS \citep{Yu2017, Bai2019}. In both works, authors used 4 consecutive ConvLSTM layers as the first part of the deep network, than \citet{Yu2017} completed the network with a fully connected layer and \citet{Bai2019} used a 3D convolutional (3DConv) layer \citep{Ji2010} followed by 2 fully connected layers.

DL has been successfully applied in different fields, gaining remarkable results. Nonetheless, a crucial issue dealing with DL remains, which is neural network architecture definition \citep{Elsken2019,xu2020curvelane,li2021autodet,ma2021scenenet}. In fact, currently employed architectures and related hyperparameters settings have mostly been developed by human experts, a manual process which is both time-consuming and error-prone. For instance, in LSTM networks many parameters and variants can be used to solve the same problem. Performance is then influenced by the final version of the LSTM networks that is used. \citet{Greff2017} called it a ``...\textit{search space odyssey}...''.  

In the Machine Learning (ML) and DL community, Bayesian Optimization (BO) \citep{Frazier2018,archetti2019bayesian} - sometimes also named Sequential Model Based Optimization (SMBO) - has recently became the standard strategy for Automated Machine Learning (AutoML) and Neural Architecture Search (NAS) \citep{Elsken2019,hutter2019automated,qazi2020survey,He2021,zoller2021benchmark}. BO is a general sample-efficient strategy for global optimization of black-boxes, which are expensive and multi-extremal functions, traditionally constrained to a box-bounded search space. Furthermore, BO has also been extended to the case of unknown constraints and/or partially defined objective functions \citep{bernardo2011optimization,hernandez2015predictive,sacher2018classification,bachoc2019gaussian,candelieri2019sequential}, as well as to the case of \emph{weakly specified search spaces} \citep{nguyen2017bayesian,nguyen2019filtering}. A recent review of automated ML techniques for DL approaches can be found in \citet{He2021}.

This work proposes the application of \textit{Bayesian Optimization} (BO) combined with \textit{Spatiotemporal-Long Short Term Memory} (ST-LSTM) network for the prediction of the steering angles in automated driving systems based on camera images (\textit{i.e.} raw pixels) taken from the front car (called BO\_ST-LSTM from now on). Unlike previous works \citep{Yu2017, Bai2019}, we have added a MaxPooling layer to reduce the complexity of the final network and we have applied the BO to optimize the hyperparameters and obtain a more suitable model for the experiment taken into account. More precisely, the main contributions of this study can be summarized as follows.\\
(1) We propose an effective and efficient ST-LSTM based network, BO\_ST-LSTM, to enhance steering wheel angle prediction. BO\_ST-LSTM has demonstrated higher accuracy with respect to competitors and a better generalization performance between validation and test set.\\
(2) We propose the use of BO as a general hyperparameter optimization framework for DL methods in the context of automated driving systems (ADS). This hybrid approach can improve the predictive power of DL architectures, more precisely ST-LSTM, by finding the most suitable configuration for the problem under study.\\
Our results can be adapted and used in the development of ADSs to help obtain a more precise prediction of steering wheel angle, leading to safer car for drivers and traffic participants.

The following sections of the paper are organized as follows. Section \ref{Sec2} will discuss related works in the field of automated driving systems (ADS). Section \ref{Sec3} will present our method of prediction of steering angles by coupling BO and ST-LSTM, while Section \ref{Sec4} will explain the experiments and performances of our method.
Finally, Section \ref{Sec5} will present discussions and conclusions of the paper.

\section{Related work} \label{Sec2}

Object recognition is an important task in different fields and it is often solved by using machine learning methods and large datasets. An example is the so-called \textit{AlexNet} \citep{Krizhevsky2012}. In this work, the authors trained a large convolutional neural networks on the subsets of ImageNet dataset \citep{Deng2009} used in the ILSVRC-2010 and ILSVRC-2010 competitions \citep{Russakovsky2015}, thereby obtaining the best results ever reported on these datasets in 2012.

In the context of ADS, \citet{Bojarski2017} proposed a neural-network-based system known as \textit{PilotNet}, which outputs steering angles given images on the road ahead. PilotNet training data contains single images sampled from videos recorder by a front-facing camera in the car, the former paired with the corresponding steering command. The PilotNet network architecture consists in a set of normalization layers, convolutional layers and fully connected layers. The proposed net is mainly tested for understanding the ability to recognize objects that can affect the steering.

More recently, \citet{Kocic2019} developed an end-to-end deep neural network (called \textit{J-net}) suitable for deployment on embedded automotive platform modifying the AlexNet \citep{Krizhevsky2012} solution, which is a convolutional neural network for image classification. Differently from AlexNet and PilotNet, J-Net is based on a set of convolutional layers and a set of max-pooling layers used to reduce the number of parameters. Compared with AlexNet and PilotNet, J-Net had comparable results in terms of predictive power, but with a lower complexity overall.

Motion planning is a fundamental technology for autonomous driving vehicles and, over the past years, novel deep learning approaches have been demonstrated to be power techniques. In \citet{Yu2017}, two main contributions can be found. Firstly, the authors introduced the Baidu Driving Dataset (DBB): DBB is a new driving dataset, which contains a 10.000-kilometer frontal camera image and a vehicle motion attitude data of real read conditions. Secondly, the authors have proposed an end-to-end reactive control model for lateral and longitudinal controls. The former referring to steering angle predictions, the latter referring to means accelerating and braking commands optimization. Similarly to other works \citep{Bojarski2017, Kocic2019}, a set of pre-processing layers, convolutional layers and fully connected layers were used for steering angle prediction. The accelerating and braking problem was solved with a Convolutional Long Short Term Memory architecture (Conv-LSTM) \citep{Shi2015}  using 5 frames, taken from the front camera car, as input. The network architecture was composed of 4 Conv-LSTM layers and a fully connected layer. 

The Conv-LSTM network combines image feature extraction capabilities from convolutional networks and the memory ability of LSTM networks. Starting from the work of \citet{Yu2017}, \citet{Bai2019} proposed the so-called Spatiotemporal-Long Short Term Memory (ST-LSTM) network for motion planning value prediction in the context of autonomous vehicle. The ST-LSTM network is composed of 4 ConvLSTM layers followed by a 3-D convolutional (3DConv) layer \citep{Ji2010} and 2 fully connected layers. As in \citet{Yu2017}, authors do not consider a single image as input but 3 continuous frames to ensure the real-time performance of motion result.

Recently, AutoML and NAS are becoming standard solutions for optimizing hyperparameters and architecture of neural networks, respectively, with examples in different application domains. For instance, in \citet{nguyen2020long} an accurate and reliable multi-step ahead prediction model based on LSTM, whose hyperparameters have been optimized through BO, has been validated on steam generator data acquired from different French nuclear power plants for prognostic and health management of the plants. In \citet{osmani2019bayesian}, an expensive and time-consuming design of a Deep Neural Network for human activity recognition has been addressed via BO in order to optimally and efficiently tune the deep neural architectures’ hyper-parameters. With respect to ADS, a recent and interesting application of BO is devoted to generate simulation scenarios in order to improve accuracy and ``safety'' of the ADS \citep{abeysirigoonawardena2019generating,zerwas2019netboa,gangopadhyay2019identification}. In \citet{Kong2020}, authors proposed a deep Q-learning (DQL)-based energy management strategy (EMS) for an electric vehicle. In this work, BO is used to optimize the hyperparameter configuration of the DQL-based EMS.

\section{Proposed method} \label{Sec3}

Firstly, the general architecture of the ADS for steering angle prediction is hereby presented, followed by a short description of the two main ADS components: BO and ST-LSTM.  

\begin{figure}[ht]
  \includegraphics[width=11cm]{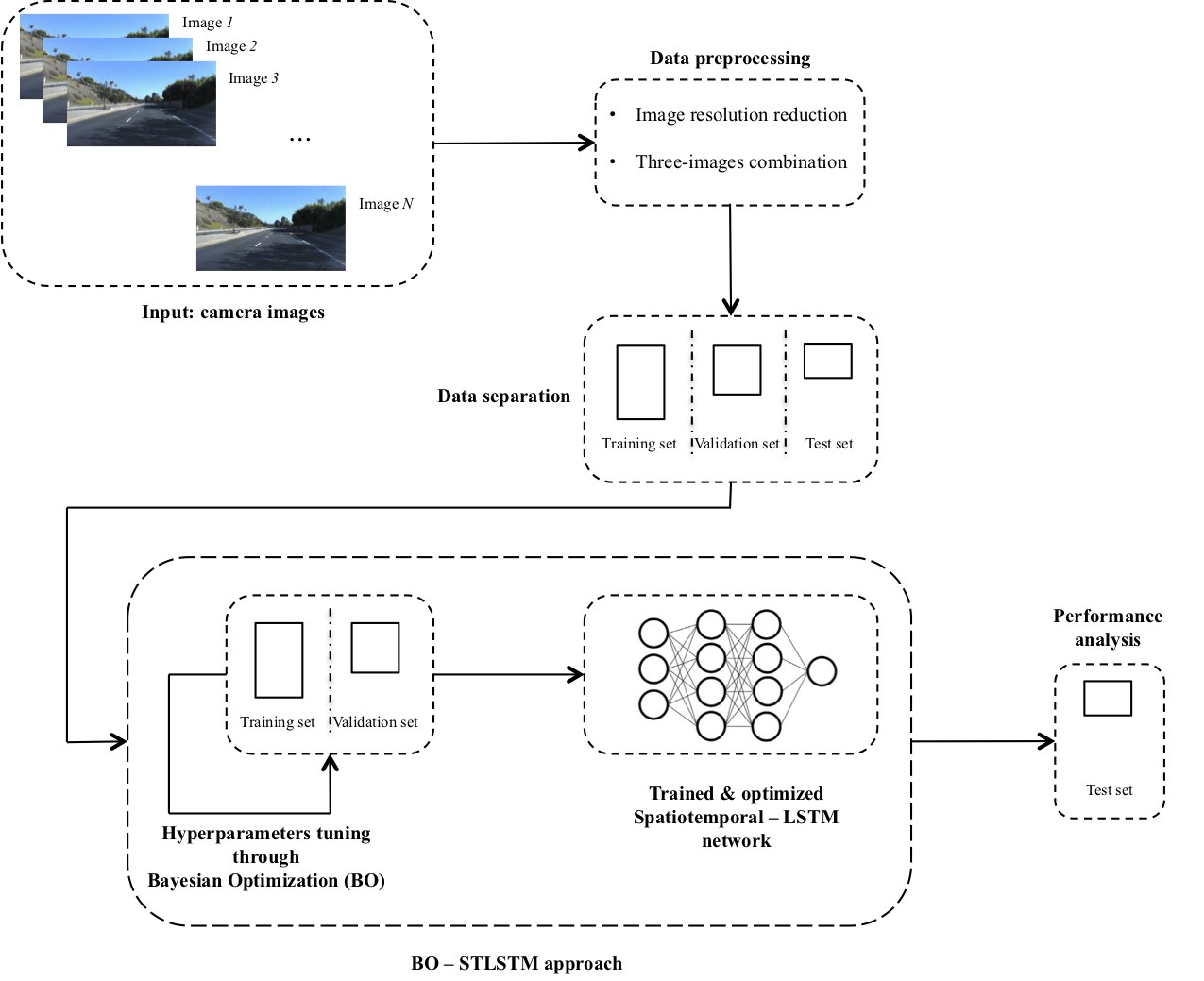}
  \centering
\caption{Overview of the BO-STLSTM system.}
\label{fig:1}       % Give a unique label
\end{figure}

Fig. \ref{fig:1} shows the general approach of the BO\_ST-LSTM system design for steering angle prediction. The first part, the data layer, is composed of two procedures, namely data preprocessing and separation. In the first procedure, several preprocessing operations are applied. In the next stage, the upper and lower sections of the images are cropped to eliminate unnecessary information; then, resolution reduction is applied to each frame for computational reasons. The size is reduced from 455 $\times$ 256 pixels to 200 $\times$ 66 pixels. This dimension is equal to the one used as input for the PilotNet \citep{Krizhevsky2012}. Finally, three consecutive images are stacked and used as the input frame dimension for the deep neural network. The steering angle associated with each tensor is the one related to the last frame. 

The second procedure, called data separation, is used to split the dataset, and a usual machine learning principle is applied. The dataset is divided into training, validation and test sets in order to learn, optimize and test the final intelligent system.

BO\_ST-LSTM layer is responsible for the optimization and learning phases of the BO\_ST-LSTM system. Training and validation sets are used to optimize the ST-LSTM network by means of Bayesian Optimization \citep{Frazier2018}. This procedure is useful to develop a more robust architecture suitable for this specific task; then, during the optimization phase, an early stopping with patience equal to 5 is used to avoid overfitting and reduce the overall computation time. Given the optimized setting, the ST-LSTM is trained and its performance is measured on the test set.

\subsection{Bayesian optimization}
Bayesian Optimization is a sample-efficient strategy for global optimization of black-boxes, expensive and multi-extremal functions, traditionally constrained to over a box-bounded search space $\Omega$:
\begin{equation}
    \min\limits_{\theta \in \Omega} g(\theta)
\end{equation}

To solve such problem (1), BO uses two key components: a \emph{probabilistic surrogate model} of the objective function $g(\theta)$ and an \emph{acquisition function} (also called \emph{infill criterion} or \emph{utility function}) that is based on the current approximation of $g(\theta)$. The optimization of the acquisition function allows to select the next promising $\theta'$ where to evaluate the objective function. The observed value, $g(\theta')$ (or $g(\theta')+\varepsilon$ in the case that the objective function is also \emph{noisy}), is then used to update the probabilistic model approximating $g(\theta)$ and the process is iterated until a given termination criteria is reached (e.g., a maximum number of function evaluations).

A Gaussian Process (GP) \citep{williams2006gaussian} is the most common choice for the probabilistic surrogate model. An alternative is offered by Random Forest (RF) \citep{Ho1995}, an ensemble learning method which, contrary to GP, is able - by construction - to deal with a complex search spaces $\Omega$, spanned by mixed, categorical and conditional components of ($\theta$). Conditional means that the value of a component of the solution vector $\theta_{[i]}$ depends on the value of at least another component $\theta_{[j]}$, with $i \neq j$.

Regardless its specific implementation, the aim of the probabilistic surrogate model is to provide an estimate (aka prediction) of $g(\theta), \forall \theta \in \Omega$, along with a measure of uncertainty about such estimate. These two elements are usually the mean and standard deviation of the prediction provided by the probabilistic surrogate model, denoted by $\mu(\theta)$ and $\sigma(\theta)$ respectively.

The acquisition function is aimed at driving the selection of the next $\theta'$ to be evaluated on the objective function, balancing between \emph{exploitation} - that is choosing $\theta'$ whose associated prediction is not worse than the best function value observed so far - and \emph{exploration} - that is choosing $\theta'$ whose prediction is largely uncertain. While exploitation is associated to \emph{local search}, exploration is associated to \emph{global search}: the first is significantly driven by $\mu(\theta)$ while the second is significantly driven by $\sigma(\theta)$.
Several acquisition functions have been proposed - an overview is provided in \citet{Frazier2018} and \citet{archetti2019bayesian} - each one offering a different mechanism to balance the exploitation-exploration trade-off. The most widely used acquisition functions are lower confidence bound (LCB), expected improvement (EI) and maximum probability of improvement (MPI).

Lower confidence bound (LCB) is an acquisition function that manages exploration—exploitation by being optimistic in the face of uncertainty:
\begin{equation}
LCB(\theta) = \mu(\theta) - \xi \sigma(\theta),
\end{equation}
where $\mu(x)$ and $\sigma(x)$ are mean value and standard deviation of the probabilistic surrogate model. $\xi \geq 0$ is the parameter to manage the trade-off between exploration and exploitation. More precisely, $\xi= 0$ is for pure exploitation; on the contrary, higher values of $\xi$ emphasizes exploration. In \citet{srinivas2012information} a schedule of $\xi$ is proposed with convergence proof.

Expected improvement (EI) measures the expectation of the improvement on $g(\theta)$ with respect to the predictive distribution of the probabilistic surrogate model.
\begin{equation}
  EI(\theta) =
    \begin{cases}
      (g(\theta^+)-\mu(\theta) - \xi) \Phi(Z)+\sigma(\theta)+\phi(Z) \quad \textrm{if} \quad \sigma(\theta) > 0 \\
      0 \quad \textrm{if} \quad \sigma(\theta) = 0
    \end{cases}       
\end{equation}

$g(\theta^+)$ is the best value of the objective function observed so far, $\xi$ is used to balance between exploration-exploitation, $\phi(Z)$ and $\Phi(Z)$ are the probability distribution and the cumulative distribution of the standardized normal, respectively, with $Z$ defined as follows: 
\begin{equation}
  Z =
    \begin{cases}
      \frac{g(\theta^+) - \mu(\theta)}{\sigma(\theta)} \quad \textrm{if} \quad \sigma(\theta) > 0 \\
      0 \quad \textrm{if} \quad \sigma(\theta) = 0
    \end{cases}       
\end{equation}

Maximum Probability of improvement (MPI) was the first acquisition function proposed in literature. As it is basically biased towards exploitation, it has been recently modified by including a parameter $\xi$ to allow for a better exploration-exploitation trade-off:
\begin{equation}
MPI(\theta) = P(g(\theta) \leq g(\theta^+) + \xi) = \Phi \bigg(\frac{g(\theta^+) + \mu(\theta) + \xi}{\sigma(\theta)}\bigg)
\label{eq:PI}      
\end{equation}

At last, selecting $\theta'$ requires to solve an auxiliary optimization problem on the same search space $\Omega$ but computationally cheaper than (1), that is minimizing $LCB(\theta)$ or maximizing $EI(\theta)$ or $MPI(\theta)$.

Denote with $D_{1:n}$ a set of initial solutions (i.e., initial design), for example sampled by using Latin Hypercube Sampling (LHS) technique. The element $D_i$ is $(\theta_i,g(\theta_i))$, with $i=1,...,n$ (i.e., we are considering the noise-free setting, without loss of generality). Then, the general BO algorithm is summarized as follows:\\

\begin{algorithm}[H]
\SetAlgoLined
 initialize $D_{1:n}$\\
 \While{$n<N$}{
  update the probabilistic surrogate model depending on $D_{1:n}$:\\
  update $\mu_n(\theta)$ and $\sigma_n(\theta)$\\
  compute the updated acquisition function $\alpha_n(\theta)$\\
  $\theta_{n+1} \leftarrow{} \underset{\theta \in \Omega}{\arg\max}$ $\alpha(\theta)$\\
  $n \leftarrow{} n+1$\\
 }
 return $\theta^+$, that is the point associated to the best value observed:\\
 $\theta^+ : g(\theta^+)=\min\big\{g(\theta_1),...,g(\theta_N)\big\}$
 \caption{Bayesian Optimization algorithm}
\end{algorithm}

\subsection{Spatiotemporal-long short term memory network}

The ST-LSTM network \citep{Bai2019} is based on Convolutional LSTM (ConvLSTM) layers, 3D Convolutional (3DConv) layers and Fully Connected (FC) layers with the aim to extract the spatiotemporal features of time-serial image segment. \citet{Bai2019} proposed architecture with 4 ConvLSTM layers combined with a batch normalization step so as to prevent the eventuality of low training efficiency caused by data distribution offset in deep networks. Furthermore, consider $N$ as the maximum number of function evaluations.

ConvLSTM uses multi-frame picture segments as input. In this way, spatial features can be extracted like a convolutional layer and the timing relationship can be obtained at the same time.

The four ConvLSTM layers  are then used to extract the temporal hidden feature information and help the model learn more effectively from the data. Following the definition of \citet{Shi2015}, all the inputs $\mathcal{X}_1, \dots, \mathcal{X}_t$, cell outputs $\mathcal{C}_1, \dots, \mathcal{C}_t$, hidden states $\mathcal{H}_1, \dots, \mathcal{H}_t$, and input and forget and output gates ($i_t, f_t, o_t$) are 3D tensors whose last two coordinates are spatial dimensions. The key equations of ConvLSTM are shown in Eq. \ref{eq:1}, where $\ast$ denotes the convolution operator and $\circ$ denotes the Hadamard product:

\begin{equation}\label{eq:1}
    \begin{aligned}
        i_t &= \sigma (W_{xi} \ast \mathcal{X}_t + W_{hi} \ast \mathcal{H}_{t-1} + W_{ci} \circ \mathcal{C}_{t-1} + b_i) \\
        f_t &= \sigma (W_{xf} \ast \mathcal{X}_t + W_{hf} \ast \mathcal{H}_{t-1} + W_{cf} \circ \mathcal{C}_{t-1} + b_f) \\
        \mathcal{C}_t &= f_t \circ \mathcal{C}_{t-1} + i_t \circ \tanh (W_{xc} \ast \mathcal{X_t} + W_{hc} \ast \mathcal{H}_{t-1} + b_c) \\
        o_t &= \sigma (W_{xo} \ast \mathcal{X}_t + W_{ho} \ast \mathcal{H}_{t-1} + W_{co} \circ \mathcal{C}_{t-1} + b_o) \\
        \mathcal{H}_t &= o_t \circ \tanh (\mathcal{C}_t)
\end{aligned} 
\end{equation}

where $\sigma$ is the activation function (\textit{i.e.} sigmoid function), $b_i, b_f, b_c$ and $b_0$ are the biases related to $i_t, f_t, \mathcal{C}_t$ and $o_t$. 

If we view the states as the hidden representations of moving objects, a ConvLSTM with a larger transitional kernel should be able to capture faster motions while one with a smaller kernel can capture slower motions \citep{Shi2015}. 

3DConv is used to better capture the temporal and spatial characteristics of videos. In fact, the traditional 2D convolutional layer operates on time-serial images using a simple convolutional layer to identify each frame of the video. Nevertheless, this method does not take into account the inter-frame motion information in the time dimension. 

The convolution operation in 3DConv has a time dimension of three, meaning that the input is composed of three consecutive frames of images. In this structure, each feature map in the convolutional layer is connected to multiple adjacent successive frames in the previous layer, thus capturing motion information.

\section{Experiments} \label{Sec4}

\begin{figure}
  \includegraphics[scale=0.8]{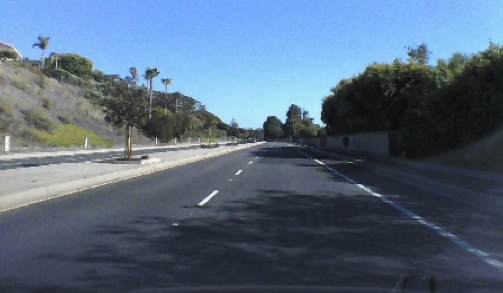}
  \centering
\caption{Image from the SullyChen Dataset \citep{Dataset}.}
\label{fig:2}       % Give a unique label
\end{figure}

The proposed methodology is evaluated and compared with the classical end-to-end driving models on the public dataset SullyChen Dataset \citep{Dataset}. The data were recorded around Rancho Palos Verdes and San Pedro, California, using a 2014 Honda Civic. A first version of the dataset has been used by \citet{Qian2018} on their work. 
Fig. \ref{fig:2} shows one example. The dataset contains both straight and mixed roads. Furthermore, pictures are taken on junctions, primary and secondary roads.

In this work, we used the updated version which includes 63.000 images of the frontal road as well as the steering angles; for computational reasons, we decided to use around the 62$\%$ (39.000) of the total number of images, of which 64$\%$ is used for training, 16$\%$ for validation and the remaining part for test purposes. Originally, images come with a dimension of $455 \times 256$ pixels; in this study, however, the size has been reduced to $200 \times 66$ pixels.

\subsection{Methodology}

Three different architectures were implemented and trained for 15 epochs, with batch size equal to 50 and validated with different setting according to their definitions: PilotNet \citep{Bojarski2017}, J-net \citep{Kocic2019} and modified version of the ST-LSTM proposed by \citet{Bai2019}. Table \ref{Table:1} summarizes settings for all networks.

\begin{table}%[width=.8\linewidth,cols=3,pos=ht]
\centering
\caption{Setting for all networks}\label{Table:1}
\begin{tabular}{c|c|c}
\hline
\textbf{PilotNet} & \textbf{J-Net} & \textbf{ST-LSTM}\\
\hline
Normalization layer & Normalization layer & Normalization layer\\
Conv2D & Conv2D & ConvLSTM2D \\
Conv2D & MaxPooling2D & BatchNormalization \\
Conv2D & Conv2D & ConvLSTM2D \\
Conv2D & MaxPooling2D & BatchNormalization \\
Conv2D & Conv2D & ConvLSTM2D \\
Flatten & MaxPooling2D & BatchNormalization \\
Dense & Flatten & ConvLSTM2D \\
Dense & Dense & BatchNormalization \\
Dense & & Conv3D \\
& & MaxPooling3D \\
& & Flatten \\
 & & Dense \\
\hline
Output value & Output value & Output value \\
\hline
\end{tabular}
\end{table}

PilotNet is a deep learning network mainly based on convolutional layers. More precisely, the network consists of 9 layers, including a normalization layer, 5 convolutional layers and 3 fully connected layers. The convolutional layers were designed to perform feature extraction and were chosen empirically through a series of experiments that varied layer configurations (see \citet{Bojarski2017} for more details). Strided convolutions were used in the first three convolutional layers with a 2 $\times$ 2 stride and a 5 $\times$ 5 kernel and a non-strided convolution with a 3×3 kernel size in the last two convolutional layers. The five convolutional layers are followed by three fully connected layers leading to an output control value (the steering angle).

As the previous network, J-Net is basically a convolutional neural network. In this case, the network consists of 5 layers, including a normalization layer, 3 convolutional layers and one fully connected layer. In order to reduce the size of the deep neural network layers, the authors applied three maxpooling operators, one after each convolutional layer. The size of all max-pooling operators is 2 $\times$ 2. Eventually, the last layer of the J-net is a fully-connected layer composed of ten nodes and followed by a simple output layer for the steering angle prediction. 

Differently from PilotNet and J-Net, ST-LSTM network is composed of a series of ConvLSTM layers and a 3DConv layer. The deep network proposed in \citet{Bai2019} consists of 6 layers, including 4 ConvLSTM layers, one 3DConv layer and one fully connected layer followed by the final output layer. Each ConvLSTM layer is followed by a batch normalization step. In order to reduce dimensionality, we introduced in this work a max-pooling layer operating of size 2 $\times$ 2 $\times$ 2 on 3D input. Furthermore, we added a dropout regularization technique before the output prediction to avoid overfitting.
This latter architecture is improved by means of  Optimization \citep{Frazier2018} to propose a more robust solution. This procedure is performed with three different acquisition functions and, in order to limit the impact of the random component, we executed 10 runs for each function. Details of this process are provided in the next sections. 

The developed architectures are trained using stochastic gradient descent. The algorithm seeks to update the weights of the model to reduce the error between actual and estimated values. 
In order to compute this error, it is necessary to choose a suitable loss function; as such, having to handle a regression problem, we resolved to use the mean squared error (MSE). As reported in Equation \ref{eq:2}, this measure is given by the average of the square of the difference between actual $y_i$ and estimated values $\hat{y}_i$.

\begin{equation}
    MSE = \displaystyle\frac{1}{n}\sum_{t=1}^{n}(y_i - \hat{y}_i)^2
\label{eq:2}
\end{equation}

\subsection{BO\_LT-STM: optimizing the hyperparameters of an ST-LSTM}
As described in Section \ref{Sec3}, Bayesian Optimization is a sample-efficient strategy for global optimization of black-boxes. BO uses probability to find a minimum of an objective function in order to obtain better performance in the testing phase and reduce the optimization time. For this specific task, we have used the mean square error on the validation set as the objective function to be minimized. Also, this procedure is executed on eight different hyper-parameters of the ST-LSTM architecture. Table \ref{Table:2} shows the parameters considered with the relative domain spaces.

\begin{table}%[width=.8\linewidth,cols=3,pos=ht]
\centering
\caption{Parameters of the ST-LSTM architecture for Bayesian Optimization}\label{Table:2}
\begin{tabular}{c|c|c}
\hline
Parameters name & Domain space & Domain type\\
\hline
ConvLSTM$_1$: Num. feature maps & $\{4, 8, 10, 16\}$ & Discrete \\
ConvLSTM$_2$: Num. feature maps & $\{4, 8, 10, 16\}$ & Discrete\\
ConvLSTM$_3$: Num. feature maps & $\{4, 8, 10, 16\}$ & Discrete\\
ConvLSTM$_4$: Num. feature maps & $\{4, 8, 10, 16\}$ & Discrete\\
Conv3D: Num. feature maps & $\{1, 2, 3\}$ & Discrete\\
FC: Num. of neurons & $\{5, 10, 25, 50\}$ & Discrete\\
Dropout & $\{0.0,0.5\}$ & Continuous\\
Learning rate (Adam optimizer) & $\{0.01, 0.001, 0.0001, 0.00001\}$ & Discrete\\
\hline
\end{tabular}
\end{table}

The BO process is performed using the Gaussian Process (GP) as a probabilistic surrogate model and through three different acquisition functions: lower confidence bound (LCB), expected improvement (EI), and maximum probability of improvement (MPI).

We compared the obtained results in order to identify the most suitable acquisition function for the optimization process. A set of initial random solutions composed of five configurations is defined to update the surrogate model at each execution. The optimization process starts from these configurations and lasts 20 iterations. 
The acquisition function selects the new configuration to be evaluated at each run by managing the trade-off between exploration and exploitation.
Therefore, in order to limit the randomness's impact in the choice of initial solutions, we have performed each optimization process with the relative acquisition function for ten times, with different seeds. 

\begin{figure}[ht]
  \includegraphics[scale=0.25]{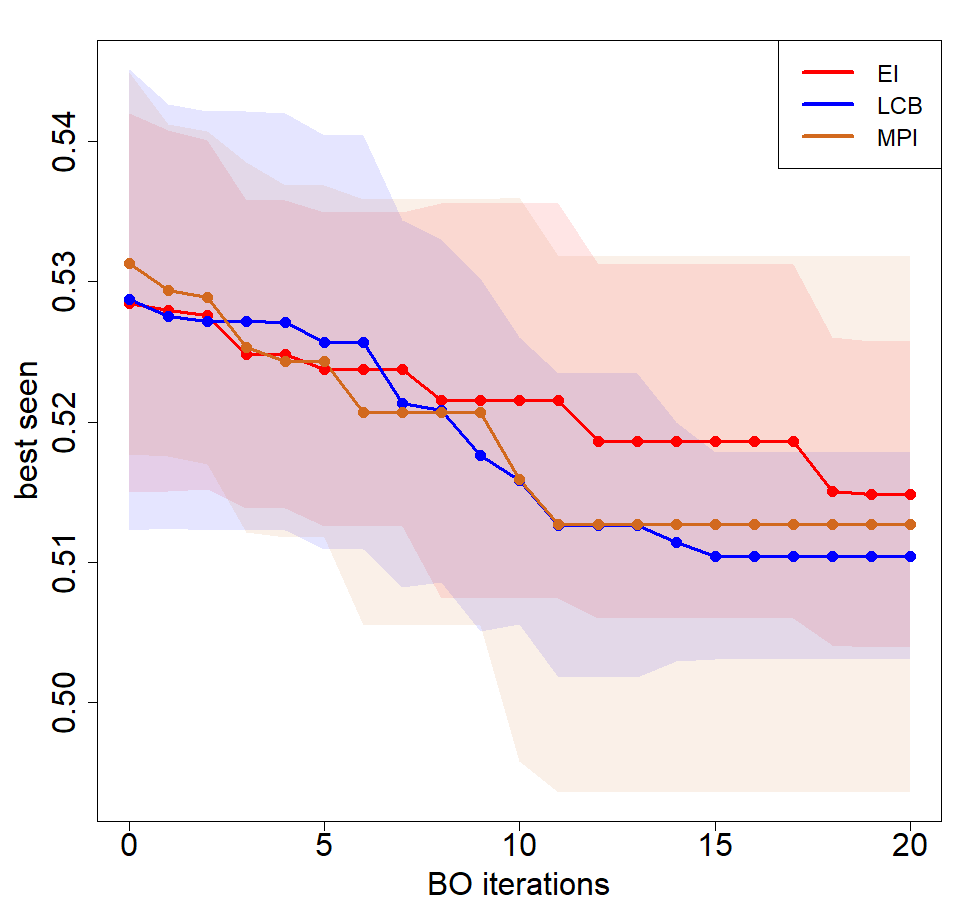}
  \centering
\caption{Comparison among GP-based BO processes using three different acquisition functions.}
\label{fig:3}       % Give a unique label
\end{figure}

Figure \ref{fig:3} shows the evolution of the so-called \textit{best seen}, that is the minimum validation error value observed during the BO iterations in this study. Solid lines represent the average, while shaded areas are standard deviation over the ten experiments. At the first iteration, the \textit{best seen} is the minimum validation error observed on the initial random configurations; then, the  \textit{best seen} of the successive 20 evaluated configurations is reported at a later stage.

\begin{figure}[ht]
  \includegraphics[scale=0.25]{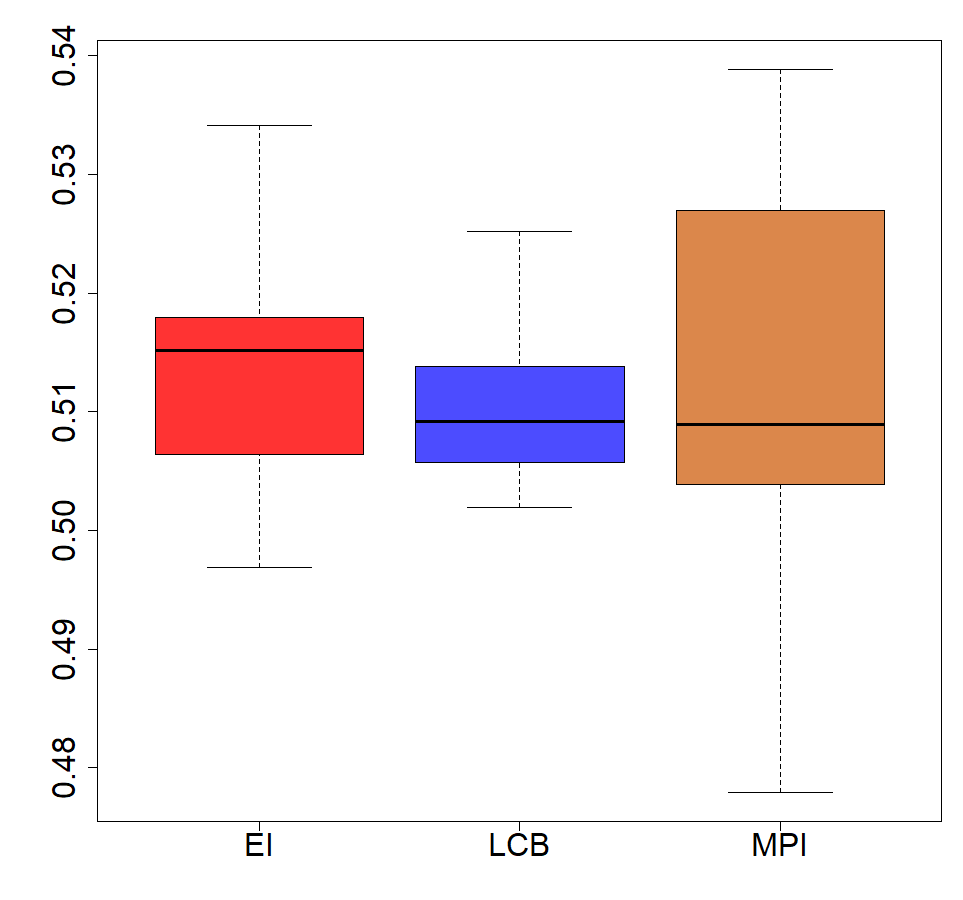}
  \centering
\caption{Comparison between the distribution of the \textit{best seen} obtained from the different acquisition function.}
\label{fig:4}       % Give a unique label
\end{figure}

All of the three acquisition functions allow, on average, to reduce the error by about two percentage points during the BO process. All of them, in fact, lead to identifying configurations with a similar error value.
Since the average results obtained are similar, the standard deviation of the various \textit{best seen} values at the end of BO was considered to identify the most performing acquisition function. As shown in Figure \ref{fig:4}, the \textit{best seen} value determined by LCB has a lower standard deviation compared to the other two acquisition functions. For this reason, ST-LSTM has been re-trained while considering the best configuration obtained through BO with the LCB acquisition function.

\begin{figure}[ht]
  \includegraphics[scale=0.19]{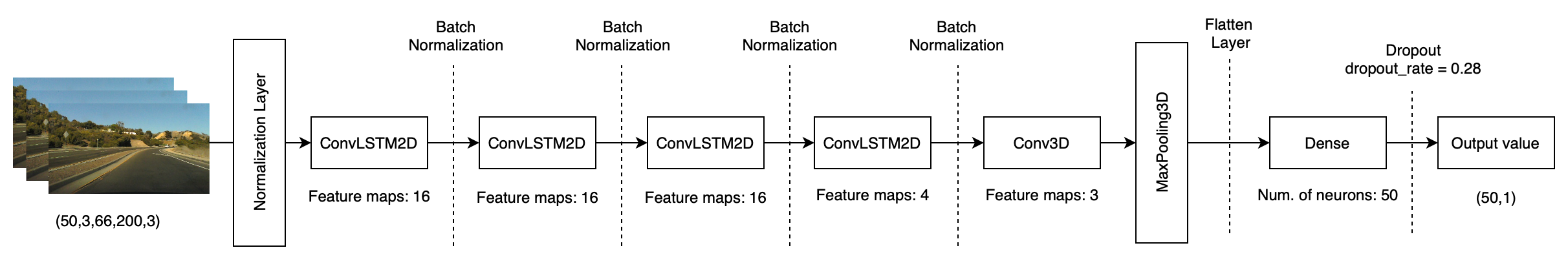}
  \centering
\caption{Structure of BO\_ST-LSTM architecture.}
\label{fig:5}       % Give a unique label
\end{figure}

Figure \ref{fig:5} displays a graphical representation of the final ST-LSTM optimized architecture. 

\subsection{Results}

This section summarizes the most relevant results of the study. The predictive performance indicators selected to compare the developed architectures are the MSE (Eq. \ref{eq:2}), the mean absolute error (MAE) and the standard deviation of absolute error (St. AE).
\begin{equation}
    AE =  \Bigl| \hat{y_i} - y_i\Bigr|, \; \; \; \; \; \; i = 1, \dots, n
\end{equation}
\begin{equation}
    MAE = \frac{1}{n} \sum_{i = 1}^{n} AE_i,
\end{equation}
\begin{equation}
    St.\;AE = \sqrt{\frac{1}{n - 1} \sum_{i = 1}^{n} \Bigl( AE_i - MAE\Bigr)^2}
\end{equation}
where $\hat{y_i}$ is the $i^{th}$ predicted value of the response system, $y_i$ is the $i^{th}$ actual steering wheel angle and $n$ is the number of observations in the evaluated set (validation or test set). Table \ref{Table:3} summarizes the overall results on training and validation sets. BO\_ST-LSTM is the best approach in mainly all the indicators on the validation set. All approaches show a possible problem related to overfitting in fact all of them obtained really good results on the training set that are not confirmed on the validation set. For this reason, the trade-off bias-variance is investigated in what follows.
\begin{table}%[width=.8\linewidth,cols=5,pos=ht]
\centering
\caption{Average values of the prediction performance indicators on the training and validation sets for the four architectures. In bold the best value for each indicator.}\label{Table:3}
\begin{tabular}{cc|cccc}
\hline
 & & \textbf{PilotNet} & \textbf{J-Net} & \textbf{ST-LSTM} & \textbf{BO\_ST-LSTM}\\
\hline
& MSE & 0.0209 & \textbf{0.0114} & 0.0405 & 0.1831 \\
Training & MAE & 0.0870	& \textbf{0.0697}	&0.1181&	0.1971\\
& St. AE & 0.1155 &	\textbf{0.0810} &	0.1630 &	0.3798\\
\hline
& MSE & 0.6814 & 0.5842 & 0.6139 & \textbf{0.5019} \\
Validation & MAE & 0.4409	&0.4262	&0.4710&	\textbf{0.4042}\\
& St. AE & 0.6979 &	0.6345 &	0.6263 &	\textbf{0.5820}\\
% MAPE & 14.0255 &	17.7187&	\textbf{8.6208} &	32.1666\\
% St. APE & 169.0831 &	877.4743 &	\textbf{45.5606}&	694.3635\\
\hline
\end{tabular}
\end{table}

Table \ref{Table:4} separately reports the MSE on training and validation, for each one of the four Deep Neural Network (DNN) models considered. Then, we have decomposed the MSE into bias and variance in order to investigate possible differences in the balance offered by the four models. More precisely, bias and variance are estimated as follows, under the noise-free assumption (i.e., if $x_i=x_j$ then $y_i=y_j$):

\begin{equation}
    Bias^2 = \Bigg(\frac{1}{N} \sum_{i=1}^N \hat y_i - \frac{1}{N} \sum_{i=1}^N y_i\Bigg)^2
\end{equation}

\begin{equation}
    Variance = \frac{1}{N} \sum_{i=1}^N \Big((y_i-\hat y_i) - \overline{err}\Big)^2
\end{equation}
with $\overline{err}=\frac{1}{N}\sum_{i=1}^N (y_i- \hat y_i)$. Finally, $MSE=Bias^2+Variance$.\\

%Each architectures is trained 20 times on 20 different training sets. Than, a point prediction is made in order to compute the expected test MSE, for a given value $x_0$, and consequently decompose it into the sum of the variance of $\hat{f}(x_0)$, the squared bias of $\hat{f}(x_0)$ and the variance of the error term $\epsilon$ as defined by Eq. \ref{eq:tradeoff}.
%\begin{equation}
%E(y_0 - \hat{f}(x_0))^2 = Var(\hat{f}(x_0)) + [Bias(\hat{f}(x_0))]^2 + %Var(\epsilon)
%\label{eq:tradeoff}
%\end{equation}

%$E(y_0 - \hat{f}(x_0))^2$ is the expected test MSE defined as the average test MSE obtained if a number of training sets are used to estimate $f$, and then tested each at $x_0$.

\begin{table}%[width=.8\linewidth,cols=7,pos=ht]
\centering
\caption{Bias-variance tradeoff decomposition.}\label{Table:4}
\begin{tabular}{c|ccc|ccc}
\hline
& & Training& & & Validation& \\
\hline
 & \textbf{MSE} & \textbf{Bias$^2$} & \textbf{Variance} & 
 \textbf{MSE} & \textbf{Bias$^2$} & \textbf{Variance}\\
\hline
PilotNet&0.0209&	0.0004&	0.0205&	0.6814&	0.0350&	0.6464 \\
J-Net&	0.0114&	0.0002&	0.0112&	0.5842&	0.0440&	0.5402\\
ST-LSTM&	0.0405&	0.0001&	0.0404&	0.6139&	0.0755&	0.5384\\
BO\_ST-LSTM&	0.1831&	0.0002&	0.1829&	0.5019&	0.0130&	0.4881  \\
\hline
\end{tabular}
\end{table}

According to such results, BO\_ST-LSTM is the most promising model, as it provides the lowest MSE on validation. This was quite expected because the goal of BO was to minimize this indicator. Moreover, BO\_ST-LSTM resulted in the highest MSE on the training set (one order of magnitude greater than the other three DNN models), making it less prone to overfitting. Finally, MSE is basically made up of the variance component for all the DNN models, both in training and validation.

Assuming that we can select only one DNN model to be deployed in a real-life application, our choice would be BO\_ST-LSTM. This conclusion is also motivated by a pairwise Mann-Whitney U test performed on the prediction errors of the four models on the validation set. More specifically, prediction error for BO\_ST-LSTM is significantly lower than the other three models ($p\text{-value}<0.001$); ST-LSTM and PilotNet are significantly similar in terms of prediction error on the validation set ($p\text{-value}=0.866$), as well as PilotNet and J-Net ($p\text{-value}=0.054$). Finally, ST-LSTM and J-Net resulted significantly different in terms of prediction error on the validation set ($p\text{-value}=0.004$).\\

All the approaches were then re-trained on the dataset consisting of both the training and validation data, setting 15 learning epochs. Results are summarized in Table \ref{Table:5}: the inclusion of validation data leads to models with small values of MSE on test, lower than MSE on validation (i.e., a reduction of around 50\%).
Results on the test set confirm that BO\_ST-LSTM is the most performing model, providing the smallest MSE again. Thus, if we could select only one DNN model, depending on the MSE on the validation set, choosing BO\_ST-LSTM would be fine. It is important to remark that, although BO\_ST-LSTM, ST-LSTM and PilotNet resulted in really close values of MSE on test set, ST-LSTM and PilotNet would be our third and fourth choice respectively.

\begin{table}%[width=.8\linewidth,cols=3,pos=ht]
\centering
\caption{MSE values on the training and test set.}\label{Table:5}
\begin{tabular}{c|cc}
\hline
 & \textbf{Training set} & \textbf{Test set} \\
\hline
PilotNet&	0.2917	&0.2810\\
J-Net	&0.0159	&0.3204\\
ST-LSTM	&0.0442	&0.2758\\
BO\_ST-LSTM	&0.0204	& \textbf{0.2700}\\

\hline
\end{tabular}
\end{table}

% \begin{table}[width=.8\linewidth,cols=2,pos=ht]
% \caption{Average values of the prediction performance indicators on the training and test set.}\label{Table:6}
% \begin{tabular*}{\tblwidth}{LL@{} LL@{} LL@{}}
% \toprule
%  & \textbf{Training set} & \textbf{Test set} \\
% \midrule
% MSE & 0.0204&	0.2700 \\
% MAE & 0.0888&	0.3848\\
% St. AE & 0.1117&	0.3492\\
% MAPE & 3.9426&	10.5435\\
% St. APE & 75.9424&	117.2471\\

% \bottomrule
% \end{tabular*}
% \end{table}

\section{Conclusion and Future Works}
\label{Sec5}

This paper is meant to address the prediction of the steering wheel angle in a self-driving system via Deep Learning. We started from an ST-LSTM, which can be considered the most suitable model for the specific application targeted, as resulted from the empirical comparison against PilotNet and J-Net. Moreover, the hyperparameters tuning of the ST-LSTM, performed via BO, allowed to decrease further prediction error of this specific DNN architecture, both on the validation and test set. More specifically, GP-based BO with LCB acquisition function proved to be the best hyperparameters optimization strategy.

Nevertheless, some limitations ought to be considered. Although BO is a sample efficient global optimization strategy, running a single hyperparameter optimization process has required - in our experimental setting - to train and evaluate 25 different ST-LSTM models (5 sampled via LHS plus 20 via BO). While this "cost" is compensated by an MSE on validation which is significantly lower than the other DNN models, the difference between the MSE of BO\_ST-LSTM and ST-LSTM, on the test set, is quite negligible. Unfortunately, a principled comparison in terms of costs is not possible because the other three DL models are from previous studies and the actual effort needed to obtain them is not quantifiable.

Future works will address (\textit{a}) the possibility to also optimize the architecture of the ST-LSTM, moving from hyperparameters optimization towards Neural Architecture Search (NAS) - and (\textit{b}) its formulation as a multi-objective or constrained optimization problem by considering not only MSE but also other requirements, such as jointly minimizing the inference time (multi-objective) or keeping it lower than a fixed threshold (constrained).

\begin{acknowledgements}
We greatly acknowledge the DEMS Data Science Lab for supporting this work by providing computational resources.
\end{acknowledgements}

\section*{Conflict of interest}
The authors declare that they have no known competing financial interests or personal relationships that could have appeared to influence the work reported in this paper.

\bibliographystyle{spbasic}      % basic style, author-year citations
\bibliography{references}   % name your BibTeX data base

%\begin{thebibliography}{}
% etc
%\end{thebibliography}

\end{document}